\documentclass[11pt,a4paper]{article}
\usepackage[hyperref]{emnlp2020}
\usepackage{times}
\usepackage{latexsym}

\usepackage{microtype}

\usepackage{amsmath, amssymb, amsthm}
\usepackage{bbm}
\usepackage{booktabs}
\usepackage{graphicx}
\usepackage{makecell}
\usepackage{hhline}
\usepackage{subcaption}

\usepackage{multirow}

\usepackage{xcolor}
\usepackage{CJKutf8}

\usepackage{pgfplotstable}
\usepackage{pgfplots}

\usepackage{url}
\usepackage[normalem]{ulem}

\usepackage{listings}

\newcommand\T{\text}

\newcommand\refsec[1]{Section~\ref{sec:#1}}

\newcommand\reffig[1]{Figure~\ref{fig:#1}}

\newcommand\reftab[1]{Table~\ref{tab:#1}}

\usepackage{color}

\newcommand{\fone}{$\T{F}_1$}
\newcommand\ph[1]{\phantom{#1}}
\newcommand{\eg}{\textit{e.g.}}

\definecolor{lblue}{HTML}{A6CEE3}
\definecolor{lgreen}{HTML}{B2DF8A}
\definecolor{lred}{HTML}{FB9A99}
\definecolor{lorange}{HTML}{FDBF6F}
\definecolor{mblue}{HTML}{80B1D3}
\definecolor{mgreen}{HTML}{B3DE69}
\definecolor{mred}{HTML}{FB8072}
\definecolor{morange}{HTML}{FDB462}
\definecolor{blue}{HTML}{1F78B4}
\definecolor{green}{HTML}{33A02C}
\definecolor{red}{HTML}{E31A1C}
\definecolor{orange}{HTML}{FF7F00}
\definecolor{dblue}{HTML}{08519C}
\definecolor{dgreen}{HTML}{006D2C}
\definecolor{dorange}{HTML}{EC7014}

\definecolor{xblue}{HTML}{4473C4}
\definecolor{xgreen}{HTML}{6FAD46}
\newcommand{\xblue}[1]{{\color{xblue} #1}}
\newcommand{\xgreen}[1]{{\color{xgreen} #1}}

\newcommand{\stanza}{Stanza}
\newcommand{\corenlp}{CoreNLP}
\newcommand{\biobert}{BioBERT}

\newcommand{\scispacy}{scispaCy}
\newcommand{\spacy}{spaCy}

\aclfinalcopy 

\title{Biomedical and Clinical English Model Packages \\in the Stanza Python NLP Library}

\author{
Yuhao Zhang\quad Yuhui Zhang \quad Peng Qi\quad \\
\textbf{Christopher D. Manning\quad Curtis P. Langlotz} \\
Stanford University\\
Stanford, CA 94305\\
  {\tt \{yuhaozhang, yuhuiz, pengqi\}@stanford.edu}\\
  {\tt \{manning, langlotz\}@stanford.edu}\\
\\
}

\date{}

\begin{document}
\maketitle

\begin{abstract}
We introduce biomedical and clinical English model packages for the \stanza{} Python NLP library \cite{qi2020stanza}.
These packages offer accurate syntactic analysis and named entity recognition capabilities for biomedical and clinical text, by combining \stanza{}'s fully neural architecture with a wide variety of open datasets as well as large-scale unsupervised biomedical and clinical text data.
We show via extensive experiments that our packages achieve syntactic analysis and named entity recognition performance that is on par with or surpasses state-of-the-art results.
We further show that these models do not compromise speed compared to existing toolkits when GPU acceleration is available, and are made easy to download and use with \stanza{}'s Python interface.
A demonstration of our packages is available at: \url{http://stanza.run/bio}.
\end{abstract}

\section{Introduction}

A large portion of biomedical knowledge and clinical communication is encoded in free-text biomedical literature or clinical notes \cite{hunter2006biomedical, jha2009use}.
The biomedical and clinical natural language processing (NLP) communities have made substantial efforts to unlock this knowledge, by building systems that are able to perform linguistic analysis \cite{mcclosky2008self, baumgartner2019craft}, extract information \cite{poon2014literome, lee2020biobert}, answer questions \cite{cao2011askhermes, jin2019pubmedqa} or understand conversations \cite{du2019extracting} from biomedical and clinical text.

\begin{figure}[t]
\centering
\includegraphics[width=.48\textwidth]{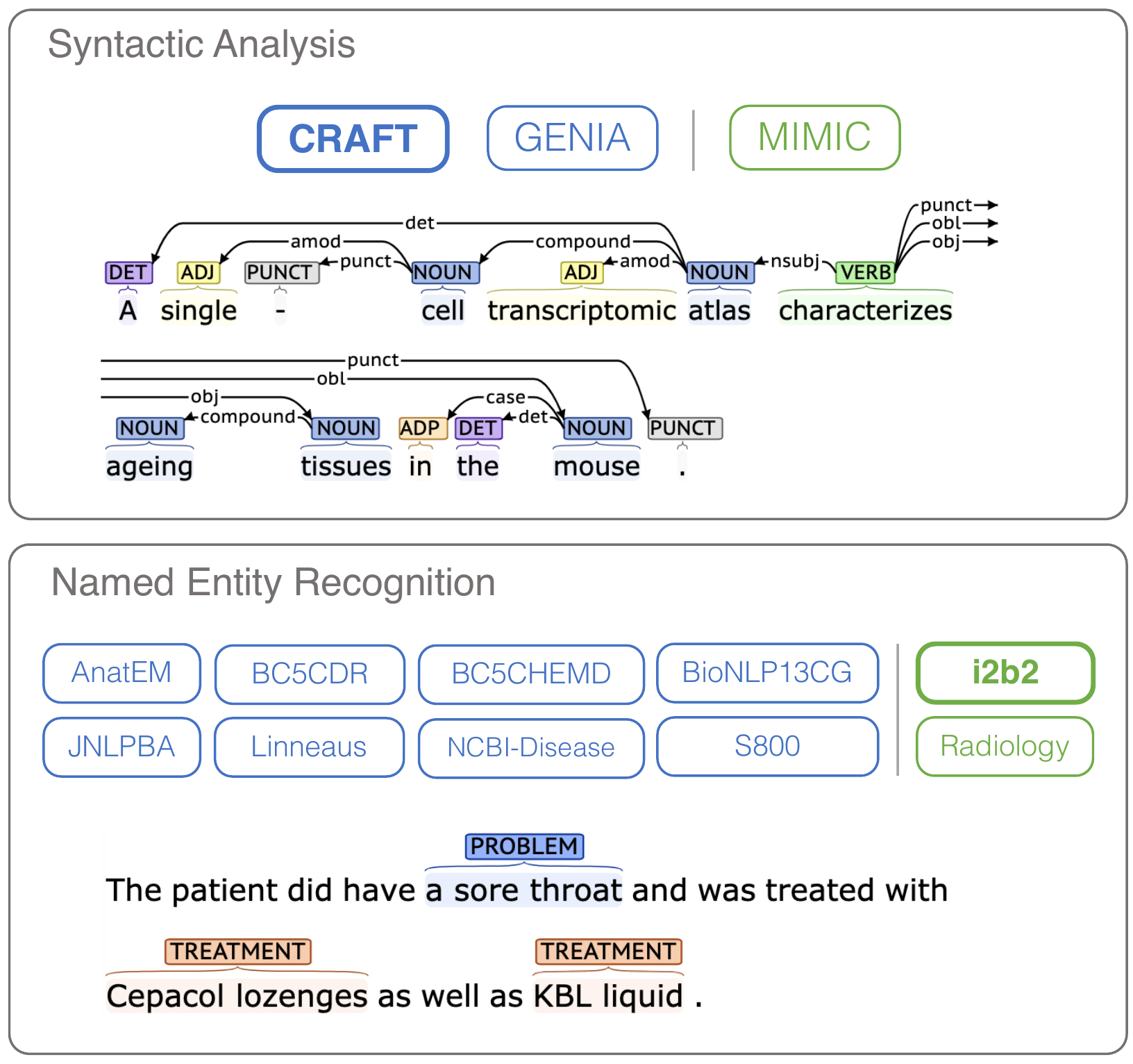}
\caption{Overview of the \xblue{\bf{biomedical}} and \xgreen{\bf{clinical}} English model packages in the Stanza NLP library.
For syntactic analysis, an example output from the CRAFT biomedical pipeline is shown; for NER, an example output from the i2b2 clinical model is shown.}
\label{fig:overview}
\end{figure}

NLP toolkits that are able to understand the linguistic structure of biomedical and clinical text and extract information from it are often used as the first step of building such systems.
While existing general-purpose NLP toolkits are optimized for high performance and ease of use, these toolkits are not easily adapted to the biomedical domain with state-of-the-art performance.
For example, the Stanford \corenlp{} library \cite{manning2014corenlp} and the \spacy{} library%
\footnote{\url{https://spacy.io/}}%
, despite being widely used by the NLP community, do not provide customized models for biomedical language processing.
The recent \scispacy{} toolkit \cite{neumann2019scispacy} extends \spacy{}'s coverage to the biomedical domain, yet does not provide state-of-the-art performance on syntactic analysis or entity recognition tasks, and does not offer models customized to clinical text processing.
On the other hand, NLP toolkits specialized for processing biomedical or clinical text, such as the MetaMap \cite{aronson2010overview} or the cTAKES \cite{savova2010mayo} libraries, often integrate sophisticated domain-specific features, yet they fall short of offering more accurate deep learning models and native Python support.

Our recently introduced \stanza{} NLP library \cite{qi2020stanza} offers state-of-the-art syntactic analysis and NER functionality with native Python support.
Its fully neural pipeline design allows us to extend its language processing capabilities to the biomedical and clinical domain.
In this paper, we present the biomedical and clinical English model packages of the \stanza{} library (\reffig{overview}).
These packages are built on top of \stanza{}'s neural system, and offer syntactic analysis support for biomedical and clinical text, including tokenization, lemmatization, part-of-speech (POS) tagging and dependency parsing, based on the Universal Dependencies v2 \cite[UDv2]{nivre2020universal} formalism; and highly accurate named entity recognition (NER) capabilities covering a wide variety of domains.

These packages include:
\begin{itemize}
  \item 2 UD-compatible biomedical syntactic analysis pipelines optimized with the publicly available CRAFT \cite{verspoor2012corpus} and GENIA \cite{mcclosky2008self} treebanks, respectively;
  \item A UD-compatible clinical syntactic analysis pipeline, trained with silver data created with clinical notes in the MIMIC-III database \cite{johnson2016mimic};
  \item 8 accurate biomedical NER models augmented with contextualized representations, achieving near state-of-the-art performance;
  \item 2 clinical NER models, including a newly introduced model specialized in recognizing entities in clinical radiology reports.
\end{itemize}

We show through extensive experiments that these packages achieve performances that are on par with or surpass state-of-the-art results.
We further show via examples and benchmarking that  these packages are easy to use and do not compromise speed, especially when GPU acceleration is available.
We hope that our packages can facilitate future research on analyzing and understanding biomedical and clinical text data.

\section{Syntactic Analysis Models}
\label{sec:ud-pipeline}

In this section, we  briefly introduce the neural syntactic analysis modules and their implementations in the Stanza toolkit, as well as how we adapt these modules to the biomedical and clinical domains.

\subsection{Modules and Implementations}

Stanza's syntactic analysis pipeline consists of modules for tokenization, sentence segmentation, part-of-speech (POS) tagging, lemmatization, and dependency parsing. 
All modules are implemented as neural network models.
We briefly introduce each component in turn and refer readers to \cite{qi2020stanza} for details.

\paragraph*{Tokenization and Sentence Splitting.}
The first step of text analysis is usually tokenization and sentence segmentation.
In Stanza, these two tasks are jointly modeled as a tagging problem over character sequences, where the model predicts whether a given character is the end of a token, a sentence, or neither.
This joint task is realized with a lightweight recurrent neural network.
We choose to combine these tasks because they are usually context-sensitive, and can benefit from joint inference to reduce ambiguity.

\paragraph*{POS Tagging.} 
Once the text is tokenized, Stanza predicts the POS tags for each word in each sentence.
We adopt a bidirectional long short-term memory network (Bi-LSTM) as the basic architecture to predict both the language-specific POS (XPOS) tags and the universal POS (UPOS) tags.
We further adapt the biaffine scoring mechanism from \citet{dozat2016deep} to condition XPOS prediction on that of UPOS, which improves the prediction consistency between XPOS and UPOS tags \cite{qi2018universal}.

\paragraph*{Lemmatization.} 
In many practical downstream applications, it is useful to recover the canonical form of a word by lemmatizing it (\eg, \emph{did}$\to$\emph{do}) for better pattern matching.
Stanza's lemmatizer is implemented as an ensemble of a dictionary-based lemmatizer and a neural sequence-to-sequence lemmatizer that operate on character sequences.
An additional classifier is built on the encoder output of the seq2seq model, to predict \emph{shortcuts} such as lowercasing and identity copy for robustness on long input sequences such as URLs.

\paragraph*{Dependency Parsing.} 
To analyze the syntactic structure of each sentence, Stanza parses it into the Universal Dependencies (UD) format \cite{nivre2020universal}, where each word in the sentence is assigned a syntactic head that is either another word in the sentence, or in the case of the root word, an artificial \emph{root} symbol.
The dependency parser in Stanza is a variant of the Bi-LSTM-based deep biaffine neural dependency parser \cite{dozat2016deep} that \citet{qi2018universal} have modified for improved accuracy.

\subsection{Biomedical Pipeline}
\label{sec:ud-pipeline-bio}

We provide two separate syntactic analysis pipelines for biomedical text by training \stanza{}'s neural syntactic pipeline on two publicly available biomedical treebanks: the CRAFT treebank \cite{verspoor2012corpus} and the GENIA treebank \cite{kim2003genia,mcclosky2008self}.
The two treebanks differ in two main ways.
First, while GENIA is collected from PubMed abstracts related to ``human'', ``blood cells'', and ``transcription factors'', CRAFT is collected from full-text articles related to the Mouse Genome Informatics database.
Second, while the CRAFT treebank tokenizes segments of hyphenated words separately (e.g., \emph{up-regulation} $\rightarrow$ \emph{up - regulation}), the GENIA treebank treats hyphenated words as single tokens.

Since both treebanks provide only Penn Treebank annotations in their original releases, to train our neural pipeline, we first convert both of them into Universal Dependencies v2 \cite{nivre2020universal} format annotations, using the Universal Dependencies converter \cite{schuster2016enhanced} in the Stanford CoreNLP library \cite{manning2014corenlp}.
To facilitate future research we make the converted files publicly available.%
\footnote{\url{https://nlp.stanford.edu/projects/stanza/bio/}}

\paragraph*{Treebank Combination.}
Since the tokenization in the CRAFT treebank is fully compatible with that in the general UD English treebanks, in practice we found it beneficial to combine the English Web Treebank (EWT) \cite{silveira14gold} with the CRAFT treebank for training the CRAFT syntactic analysis pipeline.
We show in \refsec{performance-ud} that this treebank combination improves the robustness of the resulting pipeline on both general and in-domain text.

\subsection{Clinical Pipeline}
\label{sec:ud-pipeline-clinical}

Unlike the biomedical domain, large annotated treebanks for clinical text are not publicly available.
Therefore, to build a syntactic analysis pipeline that generalizes well to the clinical domain, we create a silver-standard treebank by making use of the publicly available clinical notes in the MIMIC-III database \cite{johnson2016mimic}.
The creation of this treebank is based on two main observations.
First, we find that Stanza's neural syntactic analysis pipeline trained on general English treebanks generalizes reasonably well to well-formatted text in the clinical domain.
Second, the highly-optimized rule-based tokenizer in the Stanford CoreNLP library produces more accurate and consistent tokenization and sentence segmentation on clinical text than the neural tokenizer in Stanza trained on a single treebank.

Based on these observations, we create a silver-standard MIMIC treebank with the following procedure.
First, we randomly sample 800 clinical notes of all types from the MIMIC-III database, and stratify the notes into training/dev/test splits with 600/100/100 notes, respectively.
These numbers are chosen to create a treebank of similar size to the general English EWT treebank.
Second, we tokenize and sentence-segment the sampled notes with the default CoreNLP tokenizer.
Third, we run Stanza's general English syntactic analysis pipeline pretrained on the EWT treebank on the pre-tokenized notes, and produce syntactic annotations following the UDv2 format.
Fourth, to improve the robustness of the resulting models trained on this treebank, similar to the CRAFT pipeline, we concatenate the training split of the original EWT treebank with this silver-standard MIMIC treebank.
We show in \refsec{performance-ud} that this treebank combination again improves the robustness of the resulting pipeline on syntactic analysis tasks.

\section{Named Entity Recognition Models}

\stanza{}'s named entity recognition (NER) component is adopted from the contextualized string representation-based sequence tagger by \citet{akbik2018contextual}.
For each domain, we train a forward and a backward LSTM character-level language model (CharLM) to supplement the word representation in each sentence.
At tagging time, we concatenate the representations from these CharLMs at each word position with a word embedding, and feed the result into a standard one-layer Bi-LSTM sequence tagger with a conditional random field (CRF)-based decoder.
The pretrained CharLMs provide rich domain-specific representations that notably improve the accuracy of the NER models.

\begin{table*}[t]
  \centering
  \renewcommand{\arraystretch}{1.2}
  \begin{tabular}{llccccccc}
    \toprule
    Treebank & System & Tokens & Sents. & UPOS & XPOS & Lemmas & UAS & LAS \\
    \midrule
    CRAFT & \stanza & \bf{99.66} & \bf{99.16} & \bf{98.18} & \bf{97.95} & \bf{98.92} & \bf{91.09} & \bf{89.67}  \\
    & \corenlp & 98.80 & 98.45 & 93.65 & 96.56 & 97.99 & 83.59 & 81.81 \\
    & \scispacy & 91.49 & 97.47 & 83.81 & 89.67 & 89.39 & 79.08 & 77.74 \\
    \midrule
    GENIA & \stanza & \bf{99.81} & \bf{99.78} & \bf{98.81} & \bf{98.76} & \bf{99.58} & \bf{91.01} & \bf{89.48} \\
    & \corenlp & 98.22 & 97.20 & 93.40 & 96.98 & 97.97 & 84.75 & 83.16 \\
    & \scispacy & 98.88 & 97.18 & 89.84 & 97.55 & 97.02 & 88.15 & 86.57 \\
    \midrule
    MIMIC & \stanza & 99.18 & 97.11 & 95.64 & 95.25 & 97.37 & 85.44 & 82.81 \\
    \bottomrule
  \end{tabular}
  \caption{Neural syntactic analysis pipeline performance.
  All results are \fone{} scores produced by the 2018 UD Shared Task official evaluation script.
  All \corenlp{} (v4.0.0) and \scispacy{} (v0.2.5) results are from models retrained on the corresponding treebanks.
  UPOS results for \scispacy{} are generated by manually converting XPOS predictions to UPOS tags with the conversion script provided by \spacy.
  For \scispacy{} results \textit{scispacy-large} models are used.
  Note that the MIMIC results are based on silver-standard training and evaluation data as described in \refsec{ud-pipeline-clinical}.
  }
\label{tab:ud}
\end{table*}

\paragraph*{Biomedical NER Models.}
For the biomedical domain, we provide 8 individual NER models trained on 8 publicly available biomedical NER datasets: AnatEM \cite{pyysalo2014anatomical}, BC5CDR \cite{li2016biocreative}, BC4CHEMD \cite{krallinger2015chemdner}, BioNLP13CG \cite{pyysalo2015overview}, JNLPBA \cite{kim2004introduction}, Linnaeus \cite{gerner2010linnaeus}, NCBI-Disease \cite{dougan2014ncbi} and S800 \cite{pafilis2013species}.
These models cover a wide variety of entity types in domains ranging from anatomical analysis to genetics and cellular biology.
For training we use preprocessed versions of these datasets provided by \citet{wang2019cross}.
We include details of the entity types supported by each model in \reftab{ner}.

\paragraph*{Clinical NER Models.}
For the clinical domain, we first provide a general-purpose NER model trained on the 2010 i2b2/VA dataset \cite{uzuner20112010} that extracts problem, test and treatment entities from various types of clinical notes.
We also provide a second radiology NER model, which extracts 5 types of entities from radiology reports: anatomy, observation, anatomy modifier, observation modifier and uncertainty.
The training dataset of this NER model consists of 150 chest CT radiology reports collected from three hospitals \cite{hassanpour2016information}.
Two radiologists were trained to annotate the reports with 5 entity types with an estimate Cohen's kappa inter-annotator agreement of 0.75.
For full details of the entity types and corpora used in this dataset, we refer the readers to \citet{hassanpour2016information}.

\paragraph*{CharLM Training Corpora.}
For the biomedical NER models, we pretrain both the forward and backward CharLMs on the publicly available PubMed abstracts.
We use about half of the 2020 PubMed Baseline dump%
\footnote{ftp://ftp.ncbi.nlm.nih.gov/pubmed/baseline},
which includes about 2.1 billion tokens.
For the clinical NER models, we pretrain the CharLMs on all types of the MIMIC-III \cite{johnson2016mimic} clinical notes.
During preprocessing of these notes, we exclude sentences where at least one anonymization mask is applied (\eg, \emph{[**First Name8 (NamePattern2)**]}), to prevent the prevalence of such masks from polluting the representations learned by the CharLMs.
The final corpus for training the clinical CharLMs includes about 0.4 billion tokens.

\section{Performance Evaluation}
\label{sec:performance}

\subsection{Syntactic Analysis Performance}
\label{sec:performance-ud}

We compare \stanza's syntactic analysis performance mainly to \corenlp{} and \scispacy{}, and present the results in \reftab{ud}.
We focus on evaluating the end-to-end performance of all toolkits, i.e., each module makes predictions by taking outputs from its previous modules.
For fair comparisons, for both \corenlp{} and \scispacy, we present their results by retraining their pipelines on the corresponding treebanks using the official training scripts.
\scispacy{} results are generated by retraining the \textit{scispacy-large} models.
Notably, we find that \stanza's neural pipeline generalizes well to all treebanks we evaluate on, and achieves the best results for all components on all treebanks.

\begin{table}[t]
  \centering
  \begin{tabular}{lccc}
    \toprule
    & \multicolumn{3}{c}{CRAFT} \\
    System & XPOS & UAS & LAS \\
    \midrule
    \stanza & \bf{98.40} & \bf{93.07} & \bf{92.10} \\
    \corenlp & 97.67 & 87.75 & 86.17  \\
    \scispacy & 97.85 & 89.56 & 87.52  \\
    \bottomrule
  \end{tabular}
  \caption{Language-specific POS and UD parsing performance when gold token input is provided.
  For UAS and LAS results, gold POS tags are also provided to the parser.
  For \scispacy{} a retrained \emph{scispacy-large} model is evaluated.
  }
\label{tab:ud-gold-input}
\end{table}

\paragraph*{POS and Parsing with Gold Input.}
We notice that the much lower tokenization performance of \scispacy{} on the CRAFT treebank is due to different tokenization rules adopted:
the tokenizer in \scispacy{} is originally developed for the GENIA treebank and therefore segments hyphenated words differently from the CRAFT treebank annotations (see \refsec{ud-pipeline-bio}), leading to lower tokenization performance.
We therefore run an individual evaluation on the CRAFT treebank where gold tokenization results are provided to the POS tagger and parser at test time, and presents the results in \reftab{ud-gold-input}.
We find that even with gold tokenization input (and gold POS tags for the parser), \stanza{}'s neural pipeline still leads to substantially better performance for both POS tagging and UD parsing, with a notable gain of 5.93 and 4.58 LAS compared to \corenlp{} and \scispacy, respectively.
Our findings are in line with previous observations that neural biaffine architecture outperforms other models on biomedical syntactic analysis tasks \cite{nguyen2019pos}.

\begin{table}[t]
  \centering
    \setlength{\tabcolsep}{0.4em}
  \begin{tabular}{lccc}
    \toprule
    & \multicolumn{3}{c}{CRAFT} \\
    System & LAS & MLAS & BLEX \\
    \midrule
    \small{CRAFT-ST 2019 Baseline} & 56.68 & 44.22 & -- \\
    \small{CRAFT-ST 2019 Best} & 89.70 & 85.55 & 86.63 \\
    \midrule
    \stanza & 89.67 & 86.06 & 86.47 \\
    \bottomrule
  \end{tabular}
  \caption{Syntactic analysis performance of the \stanza{} pipeline and system submissions to the CRAFT Shared Tasks 2019 \cite{baumgartner2019craft}.
  Note that the results from \stanza{} are not directly comparable to those from the shared task, due to different dependency formalisms used (i.e., UD vs. Stanford Dependencies).
  }
\label{tab:ud-craft-st}
\end{table}

\paragraph{Comparisons to CRAFT Shared Tasks 2019 Systems.}
We further compare our end-to-end results to the state-of-the-art system in the CRAFT Shared Tasks 2019 \cite{baumgartner2019craft}, for which CRAFT is also used as the evaluation treebank.
As shown in \reftab{ud-craft-st}, we also report results for the MLAS and BLEX metrics, which, apart from dependency predictions, also take POS tags and lemma outputs into account.
We note that the results from our system are not directly comparable to those from the shared task, due to the different dependency parsing formalisms used (i.e., while we use UDv2 parse trees, the shared task used a parsing formalism similar to the Stanford Dependencies).
Nevertheless, these results suggest that the accuracy of our pipeline is on par with that of the CRAFT-ST 2019 winning system \cite{ngo2019neural}, and substantially outperforms the baseline system which uses a combination of the NLTK tokenizer \cite{bird2009natural} and the SyntaxNet neural parser \cite{andor2016globally} retrained with the CRAFT treebank.

\begin{table}[t]
  \centering
  \setlength{\tabcolsep}{0.3em}
  \begin{tabular}{lcccc}
    \toprule
     & \multicolumn{2}{c}{EWT} & \multicolumn{2}{c}{CRAFT}\\
     Training & Token \fone{} & LAS & Token \fone{} & LAS \\
     \midrule
     EWT & \bf{99.01} & \bf{83.59} & 96.09 & 68.99 \\
     CRAFT & 93.67 & 60.42 & \bf{99.66} & 89.58 \\
     Combined & 98.99 & 82.37 & \bf{99.66} & \bf{89.67} \\
     \bottomrule
  \end{tabular}
  \caption{Biomedical syntactic analysis pipeline performance. Tokenization \fone{} and LAS scores are shown for models trained with each treebank alone and a combined treebank.}
\label{tab:ud-bio}
\end{table}

\begin{table}[t]
  \centering
  \setlength{\tabcolsep}{0.3em}
  \begin{tabular}{lcccc}
    \toprule
     & \multicolumn{2}{c}{EWT} & \multicolumn{2}{c}{MIMIC}\\
     Training & Token \fone{} & LAS & Token \fone{} & LAS \\
     \midrule
     EWT & \bf{99.01} & \bf{83.59} & 92.97 & 75.97 \\
     MIMIC & 94.39 & 66.63 & 98.70 & 81.46 \\
     Combined & 98.84 & 82.57 & \bf{99.18} & \bf{82.81} \\
     \bottomrule
  \end{tabular}
  \caption{Clinical syntactic analysis pipeline performance based on a silver-standard MIMIC treebank. Tokenization \fone{} and LAS are shown for models trained with each treebank alone and a combined treebank.}
\label{tab:ud-clinical}
\end{table}

\begin{table*}[ht]
  \centering
  \begin{tabular}{lllccc}
    \toprule
     Category & Dataset & Domain & \stanza & \biobert & \scispacy \\
    \midrule
    Bio & AnatEM & Anatomy & \bf{88.18} & -- & 84.14 \\
    & BC5CDR & Chemical, Disease & \bf{88.08} & -- & 83.92 \\
    & BC4CHEMD & Chemical & 89.65 & \bf{92.36} & 84.55 \\
    & BioNLP13CG & 16 types in Cancer Genetics & \bf{84.34} & -- & 77.60 \\
    & JNLPBA & Protein, DNA, RNA, Cell line, Cell type & 76.09 & \bf{77.49} & 73.21 \\
    & Linnaeus & Species & 88.27 & \bf{88.24} & 81.74 \\
    & NCBI-Disease & Disease & 87.49 & \bf{89.71} & 81.65 \\
    & S800 & Species & \bf{76.35} & 74.06 & -- \\
    \midrule
    Clinical & i2b2 & Problem, Test, Treatment & \bf{88.13} & 86.73 & -- \\
    & Radiology & 5 types in Radiology & \bf{84.80} & -- & -- \\
    \bottomrule
  \end{tabular}
  \caption{NER performance across different datasets in the biomedical and clinical domains. 
  All scores reported are entity micro-averaged test \fone{}.
  For each dataset we also list the domain of its entity types.
  \biobert{} results are from the v1.1 models reported in \citet{lee2020biobert}; \scispacy{} results are from the \emph{scispacy-medium} models reported in \citet{neumann2019scispacy}.
  }
\label{tab:ner}
\end{table*}

\begin{table}[t]
  \centering
  \setlength{\tabcolsep}{0.2em}
  \begin{tabular}{llccc}
    \toprule
     Category & Dataset & Baseline & \stanza & $\Delta$ \\
     \midrule
     Bio & BioNLP13CG & 82.09 & 84.61 & \\
     & Linnaeus & 83.74 & 88.09 & \\
     & Average (8) & 81.90 & 84.81 & +2.91 \\
     \midrule
     Clinical & i2b2 & 86.04 & 88.08 & \\
     & Radiology & 83.01 & 84.80 & \\
     & Average (2) & 84.53 & 86.47 & +1.94 \\
     \bottomrule
  \end{tabular}
  \caption{NER performance comparison between \stanza{} and a baseline BiLSTM-CRF model without character language models pretrained on large corpora.
  For the biomedical and clinical models, an average difference over 8 models and 2 models are shown, respectively.}
\label{tab:ner-ablation}
\end{table}

\paragraph{Effects of Using Combined Treebanks.}
To evaluate the effect of using combined treebanks, we train \stanza{}'s biomedical and clinical syntactic analysis pipeline on each individual treebanks as well as the combined treebanks, and evaluate their performance on the test set of each individual treebanks.
We present the results in \reftab{ud-bio} and \reftab{ud-clinical}.
We find that by combining the biomedical or clinical treebanks with the general English EWT treebank, the resulting model is not only able to preserve its high performance on processing general-domain text, but also achieves marginally better in-domain performance compared to using the biomedical and clinical treebanks alone.
For example, while pipeline trained on the EWT treebank alone is only able to achieve 68.99 LAS score on the CRAFT test set, pipeline trained on the combined dataset achieves the overall best 89.57 LAS score on the CRAFT test set, with only 1.2 LAS drop on the EWT test set.
This suggests that compared to using the in-domain treebank alone, using the combined treebanks improves the robustness of \stanza{}'s pipeline on both in-domain and general English text.

\subsection{NER Performance}
\label{sec:performance-ner}

We compare \stanza{}'s NER performance to \biobert{}, which achieves state-of-the-art performance on most of the datasets tested, and \scispacy in \reftab{ner}.
For both toolkits we compare to their official reported results \citep{lee2020biobert, neumann2019scispacy}.
We find that on most datasets tested, \stanza{}'s NER performance is on par with or superior to the strong performance achieved by \biobert, despite using considerably more compact models.
Substantial difference is observed on the BC4CHEMD and NCBI-Disease datasets, where \biobert{} leads by 2.71 and 2.22 in \fone, respectively, and on the S800 dataset, where \stanza{} leads by 2.29 in \fone.
Compared to \scispacy{}, \stanza{} achieves substantially higher performance on all datasets tested.
On the newly introduced Radiology dataset, \stanza{} achieves an overall \fone{} of 84.80 on five entity types.

\paragraph*{Effects of Using Pretrained Character LMs.}
To understand the effect of using the domain-specific pretrained CharLMs in NER models, on each dataset we also trained a baseline NER model where the pretrained LM is replaced by a randomly initialized character-level BiLSTM, which is fine-tuned with other components during training.
We compare \stanza's full NER performance with this baseline model in \reftab{ner-ablation}.
We find that by pretraining \stanza{}'s CharLMs on large corpora, we are able to achieve an notable average gain of 2.91 and 1.94 \fone{} on the biomedical and clinical NER datasets, respectively.

\begin{table}
  \small
  \centering
  \setlength{\tabcolsep}{0.4em}
  \newcommand{\pha}{\ph{$^\ast$}}
  \begin{tabular}{lccccccc}
    \toprule
    \multirow{2}{1cm}{Task} & \multicolumn{2}{c}{\stanza} && \corenlp && \multicolumn{2}{c}{\biobert}\\
    \cline{2-3}\cline{5-5}\cline{7-8}
    & \textsc{cpu} & \textsc{gpu} && \textsc{cpu} && \textsc{cpu} & \textsc{gpu}\\
    \midrule 
    Syntactic & 6.83$\times$ & \bf{1.42}$\times$ && 7.23$\times$ && -- & -- \\
    NER & 14.8$\times$ & \bf{0.95}$\times$ && -- && 121$\times$ & 4.59$\times$ \\
    \bottomrule
  \end{tabular}
  \caption{Annotation runtime of various toolkits relative to \scispacy{} (CPU) on the CRAFT biomedical treebank and JNLPBA NER test sets.
  For each toolkit, an average runtime over three independent runs on the same machine are reported.
  For reference, on the compared syntactic and NER tasks, \scispacy{} is able to process 6909 and 5415 tokens per second, respectively.}
\label{tab:speed}
\end{table}

\subsection{Speed Comparisons}

We compare the speed of \stanza{} to \corenlp{} and \scispacy{} on syntactic analysis tasks, and to \scispacy{} and \biobert{}%
\footnote{For \biobert{} we implemented our own code to run inference on the test data, since an inference API is not provided in the \biobert{} official repository.}%
on the NER task.
We use the CRAFT test set, which contains about 1.2 million raw characters, for benchmarking the syntactic analysis pipeline, and the test split of the JNLPBA NER dataset, which contains about 101k tokens, for benchmarking the NER task.
Apart from CPU speed, we also benchmark a toolkit's speed on GPU whenever GPU acceleration is available.
Experiments are run on a machine with 2 Intel Xeon Gold 5222 CPUs (14 cores in total).
For GPU tests a single NVIDIA Titan RTX card is used.

We show the annotation runtime of each toolkit relative to \scispacy{} in \reftab{speed}.
We find that for syntactic analysis, \stanza's speed is on par with \scispacy{} when GPU is used, although it is much slower when only CPU is available.
For NER, with GPU acceleration \stanza{}'s biomedical models are marginally faster than \scispacy{}, and are considerably faster than \biobert{} in both settings, making them much more practically useful.

\section{System Usage and Examples}

We provide a fully unified Python interface for using \stanza{}'s biomedical/clinical models and general NLP models.
The biomedical and clinical syntactic analysis pipelines can be specified with a \emph{package} keyword.
The following minimal example demonstrates how to download the CRAFT biomedical package and run syntactic analysis for an example sentence:

\begin{lstlisting}
import stanza
# download CRAFT syntactic analysis package
stanza.download('en', package='craft')
# initialize pipeline
nlp = stanza.Pipeline('en', package='craft')
# annotate biomedical text
doc = nlp('A single-cell transcriptomic atlas characterizes ageing tissues in the mouse.')
# print out dependency tree
doc.sentences[0].print_dependencies()
\end{lstlisting}

For NER, \stanza{}'s biomedical and clinical models can be specified with the \emph{processors} keyword.
The following minimal example demonstrates how to download the i2b2 clinical NER model along with the MIMIC clinical pipeline, and run NER annotation over an example clinical text:

\begin{lstlisting}
import stanza
stanza.download('en', package='mimic', processors={'ner': 'i2b2'})
# initialize pipeline
nlp = stanza.Pipeline('en', package='mimic', processors={'ner': 'i2b2'})
# annotate clinical text
doc = nlp('The patient had a sore throat and was treated with Cepacol lozenges.')
# print out all entities
for ent in doc.entities:
    print(f'{ent.text}\t{ent.type}')
\end{lstlisting}

To easily integrate with external tokenization libraries, \stanza{}'s biomedical and clinical pipelines also support annotating text that is pre-tokenized and sentence-segmented.
This can be easily specified with a \emph{tokenize\_pretokenized} keyword.
The following example demonstrates how to run NER annotation over two pre-tokenized sentences passed into the pipeline as Python lists:

\begin{lstlisting}
nlp = stanza.Pipeline('en', package='mimic', processors={'ner': 'i2b2'}, tokenize_pretokenized=True)
# annotate pre-tokenized sentences
doc = nlp([['He', 'had', 'a', 'sore', 'throat', '.'], ['He', 'was', 'treated', 'with', 'Cepacol', 'lozenges', '.']])
doc.sentences[0].print_dependencies()
\end{lstlisting}

For full details on how to use the biomedical and clinical models, please see the \stanza{} website: \url{https://stanfordnlp.github.io/stanza/}.

\section{Conclusion}

We presented the biomedical and clinical model packages in the \stanza{} Python NLP toolkit.
\stanza{}'s biomedical and clinical packages offer highly accurate syntactic analysis and named entity recognition capabilities, while maintaining competitive speed with existing toolkits, especially when GPU acceleration is available.
These packages are highly integrated with \stanza's existing Python NLP interface, and require no additional effort to use.
These packages are going to be continuously maintained and expanded as new resources become available.

\bibliography{main}
\bibliographystyle{acl_natbib}

\end{document}